\pdfoutput=1
\documentclass[11pt]{article}
\usepackage[preprint]{acl}
\usepackage{times}
\usepackage{latexsym}
\usepackage[T1]{fontenc}
\usepackage[utf8]{inputenc}
\usepackage{microtype}
\usepackage{inconsolata}
\usepackage{graphicx}
\usepackage{hyperref}
\usepackage{url}
\usepackage{booktabs}
\usepackage{amsmath}        % math
\usepackage{amssymb}        % math symbols
\usepackage{amsfonts}       % blackboard math symbols
\usepackage{bm}
\usepackage[capitalise,noabbrev]{cleveref} % better referencing
\usepackage{xcolor} % Required for defining colors
\definecolor{Gray}{gray}{0.9} % Custom gray color from your example
\definecolor{lightyellow}{rgb}{1,1,0.8} % For highlighting, from your example
\definecolor{lightred}{rgb}{1.0, 0.8, 0.8} % For highlighting, from your example
\definecolor{lightgreen}{rgb}{0.8,1,0.8} % For highlighting, from your example
\definecolor{promptbg}{RGB}{240,240,240} % For llmprompt, from your example
\definecolor{promptframe}{RGB}{200,200,200} % For llmprompt, from your example

\usepackage{tcolorbox}
\tcbuselibrary{breakable} % Allows tcolorboxes to break across pages
\tcbuselibrary{listings} % For including listings directly in tcolorbox (optional if using lstlisting env)

\usepackage{listings} % For lstlisting environment (used in your appendix)
\usepackage{enumitem} % For customizing lists (e.g., nosep)
\usepackage{soul}     % For highlighting text (\hl), from your example
\usepackage{framed}   % For llmprompt, from your example
\usepackage{caption}  % For \captionof if needed, and general caption styling

\usepackage{tabularx, boldline, adjustbox}
\usepackage{array}
\usepackage{colortbl}
\usepackage{multirow}
\usepackage{algorithm, algpseudocode} % If you have algorithms

\newtcolorbox{appendixartifact}[2][]{%
  colback=gray!10!white,
  colframe=gray!50!black,
  breakable,
  width=\textwidth,
  title=#2,
  fonttitle=\bfseries,
  boxrule=0.5mm,
  arc=2mm,
  #1 % Allows for additional tcolorbox options
}

\lstdefinestyle{mypromptstyle}{
  basicstyle=\ttfamily\small,
  breaklines=true,
  columns=flexible,
  keepspaces=true, % Important for prompts
  showstringspaces=false,
  xleftmargin=2mm
}

\lstdefinestyle{mycodestyle}{
  basicstyle=\ttfamily\small,
  breaklines=true,
  columns=flexible,
  showstringspaces=false,
  xleftmargin=2mm
}

\author{Philip Lippmann \\
  Delft University of Technology \\
  \texttt{p.lippmann@tudelft.nl} \\\And
  Jie Yang \\
  Delft University of Technology \\
  \texttt{j.yang-3@tudelft.nl}}
\newcommand{\ModelName}{ZEST} 
\newcommand{\Dsynth}{\mathcal{D}_{\text{synth}}}
\newcommand{\Dex}{\mathcal{D}_{\text{ex}}}
\usepackage{amsmath,amsfonts,bm}

\def\eqref#1{equation~\ref{#1}}
\def\1{\bm{1}}

\def\mC{{\bm{C}}}

\DeclareMathAlphabet{\mathsfit}{\encodingdefault}{\sfdefault}{m}{sl}
\SetMathAlphabet{\mathsfit}{bold}{\encodingdefault}{\sfdefault}{bx}{n}

\title{Zero-Shot Contextual Embeddings via Offline Synthetic Corpus Generation}

\begin{document}

\maketitle

\begin{abstract}
Context-aware embedding methods boost retrieval accuracy by conditioning on corpus statistics (e.g., term co-occurrence and topical patterns) extracted from neighboring documents. 
However, this context-aware approach requires access to the target corpus or requires domain-specific finetuning, posing practical barriers in privacy-sensitive or resource-constrained settings.
We present \ModelName{}, a zero-shot contextual adaptation framework that replaces real corpus access with a one-time offline synthesis of a compact proxy. 
Given only a handful exemplar documents representative of the general target domain, we use a multi-step hierarchical procedure to generate a synthetic context corpus of several hundred documents that aims to emulate key domain-specific distributions.
At inference, the frozen context-aware encoder uses this proxy corpus -- without any finetuning or target corpus access -- to produce domain-adapted embeddings.
Across the MTEB benchmark, \ModelName{}'s zero-shot synthetic context adaptation using only five example documents performs within 0.5\% of models leveraging full target corpus access -- demonstrating remarkable efficacy without any retraining.
\ModelName{} thus provides a practical method for deploying high-performance, adaptable embeddings in constrained environments. 
\end{abstract}

%%%%%%%%%%%%%%%%%%%%%%%%%%%%%%%%%%%%%%%%%%%%%%%%%%%%%%%%

\begin{figure*}[t!]
    \centering
    \includegraphics[width=\textwidth]{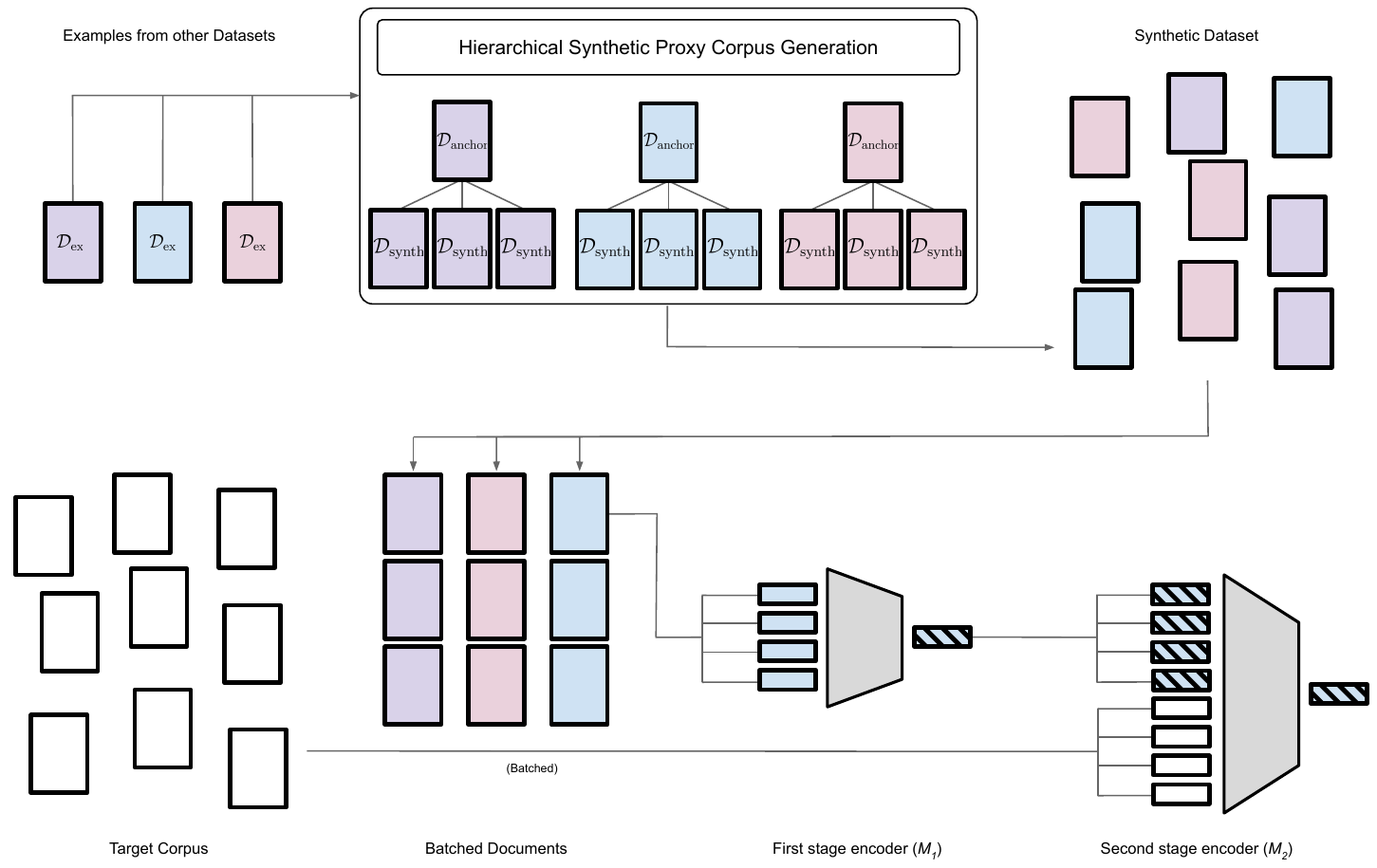} 
    \caption{Overview of the \ModelName{} framework for zero-shot contextual adaptation.
    \textbf{Offline Synthesis (Top):} From a few example documents ($\Dex$), domain anchors ($\mathcal{D}_{\text{anchor}}$) are sequentially generated to ensure thematic diversity. These anchors then guide the parallel generation of a synthetic proxy corpus $\Dsynth$.
    \textbf{Online Inference (Bottom):} A frozen, pretrained context-aware model ($M_1, M_2$) embeds new documents or queries ($d/q$) by conditioning on $\Dsynth$ (with its document representations pre-processed via $M_1$) instead of the inaccessible target corpus $\mathcal{D}$. This provides domain-adapted embeddings without requiring model retraining or direct corpus access.}
    \label{fig:overview}
\end{figure*}

%%%%%%%%%%%%%%%%%%%%%%%%%%%%%%%%%%%%%%%%%%%%%%%%%

\section{Introduction}

Effective neural information retrieval relies heavily on high-quality dense vector representations for documents and queries~\citep{reimers-2019-sentence-bert, karpukhin2020dpr}. However, standard embedding approaches typically generate these representations without dynamically incorporating information from the specific target corpus being searched~\citep{ni2021gtr}. This lack of dynamic context sensitivity limits their adaptability and retrieval performance, especially when deployed in domains that differ from their pretraining data~\citep{thakur2021beir}, a limitation partially addressed by traditional methods leveraging corpus statistics~\citep{robertson2009probabilistic}. However, such lexical methods based on surface counts have largely been superseded, as their bag-of-words nature fundamentally fails to capture semantic meaning.% or complex relational patterns that are crucial for nuanced retrieval and effectively modelled by modern neural techniques.

To provide neural embeddings with sensitivity to the target corpus, context-aware architectures have recently emerged~\citep{morris2024contextual}. Such models utilize multi-stage processing, where the final embedding is conditioned on representations derived from neighboring documents within the target corpus. Although these methods significantly enhance retrieval by tailoring embeddings to domain-specific characteristics, their requirement to access the target corpus during inference is a critical limitation. Practical constraints related to data privacy or corpus scale often make such corpus access infeasible; for instance, sensitive medical documents may not be exposed during inference. %Consequently, context-aware models revert to a contextless mode, negating their primary advantage.

This critical limitation motivates the search for alternative ways to provide essential domain signals to context-aware models. While large language models (LLMs) possess good generative capabilities for retrieval data~\citep{shao2025reasonirtrainingretrieversreasoning}, their effective use in this zero-access environment is non-trivial. The core challenge is not merely generating domain-relevant text, but ensuring that any LLM-derived output possesses an appropriate degree of representational fidelity.

In this work, to overcome this critical barrier, we introduce \underline{z}ero-shot \underline{e}mbeddings via \underline{s}ynthetic contex\underline{t} (\ModelName{}), enabling effective contextual adaptation without accessing the target corpus. 
\ModelName{}, depicted in \Cref{fig:overview}, operates in two phases. 
First, during \emph{offline synthesis}, a LLM is employed to hierarchically generate a compact set of domain anchors from a minimal set of example documents randomly chosen from a domain-relevant source. 
The example documents typify the target domain’s characteristics but remain distinct from the actual target corpus data. 
Using the domain anchors, a synthetic proxy corpus is generated that approximates the actual distribution of the target corpus. 
Second, during \emph{online inference}, the generated synthetic corpus serves as the contextual input to a pretrained, unmodified context-aware embedding architecture. 
Thus, \ModelName{} computes context-influenced embeddings, effectively simulating domain adaptation without requiring direct corpus access or any parameter updates.

Our thorough validation on relevant retrieval benchmarks demonstrates that \ModelName{} -- using only $k=5$ example documents -- significantly outperforms strong context-agnostic baselines and achieves performance comparable to context-aware models that utilize full corpus access. Although synthesizing a corpus of, in our case, a few hundred documents involves upfront computational resources, this process is performed once per domain, offline, making it practical and cost-effective for real-world deployments. \ModelName{} thus provides a viable and efficient strategy for deploying adaptable document embeddings in environments constrained by corpus access due to concerns regarding privacy or scale.

%%%%%%%%%%%%%%%%%%%%%%%%%%%%%%%%%%%%%%%%%%%%%%%%%
\section{Background}
\label{sec:background}

This section first revisits dense neural retrieval, then describes the class of context-aware embedding architectures most relevant to our work, and finally formalizes the zero-shot deployment setting that \ModelName{} is designed to
address.

\paragraph{Dense retrieval.}
Given a query $q$ and a corpus $\mathcal{D}$, dense vector retrieval methods learn two neural encoders
$\phi:D \rightarrow \mathbb{R}^{n}$ 
and
$\psi:Q \rightarrow \mathbb{R}^{n}$
that map documents and queries into a shared vector space in which relevance is scored with the dot product
$f(d,q)=\phi(d) \cdot \psi(q)$.
Training typically relies on a contrastive loss that favours high similarity
for relevant pairs $(q,d^{+})$ and low similarity for negatives
$(q,d^{-})$:
\begin{equation}
\mathcal{L}
   = -\log
     \frac{\exp \bigl(f(d^{+},q)/\tau\bigr)}
          {\displaystyle\sum_{d'\,\in\,\mathcal{N}(q)}
           \exp \bigl(f(d',q)/\tau\bigr)} ,
\label{eq:contrastive_loss_background}
\end{equation}
where $\mathcal{N}(q)$ is a set of negatives~\citep{robinson2021contrastivelearninghardnegative} and $\tau$ a temperature
hyper-parameter.
%Most information retrieval work, including ours, omits the positive example from the denominator, whereas the InfoNCE variant places it inside the normaliser; both options are valid but lead to slightly different gradients.
Although these biencoders are highly effective, their embeddings remain
\emph{context-agnostic} and therefore sensitive to corpus shift
\citep{thakur2021beir}.

\paragraph{Context-aware embedding architectures.}
Recent retrievers mitigate the limitation of context-agnostic embeddings by incorporating information from the target corpus at inference time. These approaches vary in their mechanism. One prominent strategy, which our work adapts for zero-shot settings, involves a multi-stage architecture exemplified by Contextual Document Embeddings (CDE)~\citep{morris2024contextual}. Here, a first-stage encoder $M_{1}$ processes a sample of $J$ context documents $\{d_{1},\dots,d_{J}\}\subset D$ from the target corpus. A second-stage encoder $M_{2}$ then conditions the final embedding of a target text (document $d$ or query $q$) on these context representations alongside the text's own token sequence.
Formally, the first-stage encoder $M_1$ maps an input document to a vector in $\mathbb{R}^h$. The second-stage encoder is $M_{2}:\mathbb{R}^{J\times h}\times\mathbb{R}^{T\times e} \rightarrow \mathbb{R}^{n}$,
so that, given a document $d$ whose tokens are embedded as
$E(d)=\bigl\{E(w_{1}),\dots,E(w_{T})\bigr\}$,
the contextual document representation is a single dense vector:
\begin{equation}
\phi\bigl(d;\mathcal{D}\bigr)=
      M_{2} \bigl(M_{1}(d_{1}),\dots,M_{1}(d_{J}),E(d)\bigr).
\label{eq:cde_phi}
\end{equation}
The query counterpart $\psi(q;D)$ is computed similarly. This approach directly modifies the final embedding vector based on corpus context before standard similarity search (e.g., dot product).

This differs from unsupervised adaptation methods such as GPL~\citep{wang2022gpl} and Boot\&Switch~\citep{jiang2023bootswitch}, which require corpus access or additional tuning. It also contrasts with late-interaction systems such as ColBERT~\citep{khattab2020colbert}, which, while domain-robust, incur higher online computational costs due to their token-level interaction mechanisms at query time, unlike the single-vector representations produced by CDE.

By explicitly exposing corpus statistics (such as term frequency or topical patterns via neighbor documents) to the embedding generation process, CDE-style models improve robustness across domains. However, their reliance on accessing the target corpus $\mathcal{D}$ at inference time remains a significant practical hurdle. Our work focuses on overcoming the corpus access requirement. % specifically for CDE-style architectures.

\paragraph{Problem setting: zero-shot contextual adaptation.}
We consider deploying a \emph{frozen} context-aware model $(M_{1},M_{2})$ in a
new domain where the full corpus $\mathcal{D}$ is inaccessible. %, for example due to privacy or scale constraints.
Instead, in lieu of supplying documents from the target corpus, the practitioner can supply only a small exemplar set
$\mathcal{D}_{\text{ex}}=\{d^1_{\text{ex}},\dots,d^k_{\text{ex}}\}$ that typifies the general domain.
The challenge is to generate domain-sensitive embeddings
$\phi(d; \cdot)$ and
$\psi(q; \cdot)$ without accessing $\mathcal{D}$ and without any
parameter updates. The remainder of this paper presents \ModelName{}, a solution to this problem.

%%%%%%%%%%%%%%%%%%%%%%%%%%%%%%%%%

\section{Method: Zero-Shot Embeddings via Synthetic Context}
\label{sec:method}  

\ModelName{} enables corpus-aware adaptation without direct corpus access by substituting the real neighbor set required by context-aware models with a compact, LLM-generated proxy corpus $\Dsynth$. This involves a one-time offline synthesis phase and an online inference phase using the fixed proxy.

\subsection{Rationale: Leveraging Synthetic Context}
Context-aware retrievers rely on neighbor documents to capture domain statistics (\cref{sec:background}). Our hypothesis is that an LLM, guided by few exemplars $\Dex$, can generate a synthetic corpus $\Dsynth$ whose statistics sufficiently approximate the target domain's regularities. If the frozen model’s first-stage encoder $M_{1}$ is applied to the members of $\Dsynth$ and the resulting vectors are fed into the second stage $M_{2}$ together with the query or document to be embedded, then the outputs $\phi(d;\Dsynth)$ and $\psi(q;\Dsynth)$ should exhibit much of the desired domain adaptation -- achieving effective zero-shot contextualization without $\mathcal{D}$.

\subsection{Offline Phase: Few-Shot Synthetic-Context Generation}
\label{sec:gen}

\paragraph{Input: Domain-specific examples.}
The offline pipeline begins with a small exemplar set of $k$ documents, $\mathcal{D}_{\text{ex}}=\{d_{\text{ex}}^{1},\ldots,d_{\text{ex}}^{k}\}$, selected to typify the target domain (for instance, finance or healthcare) which are sourced separately from $\mathcal{D}$. These serve as concrete stylistic and topical anchors for the LLM. While these examples provide crucial domain signals, the specific selection is not expected to be overly sensitive, particularly because the target corpus $\mathcal{D}$ is inaccessible by design, which makes fine-grained optimization of the exemplar set infeasible. As such, the primary goal is to provide the LLM with a general sense of the domain's characteristics, rather than perfectly matching unknown corpus specifics.% We find that the quality of the LLM itself plays a more dominant role in the quality of $\Dsynth$ (see \cref{sec:ablation}).

\paragraph{Hierarchical Synthetic Corpus Generation via Domain Anchors.}
To enhance the representational fidelity and thematic coherence of the synthetic context, we introduce a hierarchical generation approach based on explicit \emph{domain anchors}. We hypothesize that this intermediate anchor step is beneficial because it (i) explicitly encourages topical diversity across the final synthetic corpus, preventing fixation on only a few aspects of the initial exemplars, (ii) mitigates potential mode collapse where the LLM might over-produce content related to a single dominant theme, and (iii) grants finer-grained semantic control when expanding each focused anchor into multiple full documents. Concretely, this procedure unfolds in two steps.

\emph{\textbf{Step 1: Domain Anchor Generation.}} To establish a diverse set of thematic seeds that broadly represent the target domain, the LLM is prompted to \textit{sequentially} generate $A$ domain anchor documents, $\mathcal{D}_{\text{anchor}} = \{a_1, \ldots, a_A\}$, from the exemplar set $\Dex$. Each anchor $a_i$ is a concise text capturing a distinct topical or stylistic facet observed in $\Dex$. By generating anchors one after another, the process can be guided to ensure each new anchor explores different characteristics of the exemplars, thereby constructing a varied foundation for the subsequent corpus expansion. Practically, these anchors are generated by instructing the LLM to produce brief documents that explicitly highlight key concepts, terminology, and typical stylistic attributes of the domain as evidenced in $\Dex$.

\emph{\textbf{Step 2: Synthetic Corpus Expansion.}} Next, the complete synthetic corpus $\Dsynth$ is created by expanding upon these domain anchors. For each anchor document $a_i \in \mathcal{D}_{\text{anchor}}$, the LLM generates a corresponding subset of synthetic documents. This generation for each anchor can proceed in \textit{parallel}, with the LLM prompted to elaborate on and diversify the theme encapsulated by $a_i$. This ``branching out'' from each anchor aims to populate $\Dsynth$ with a rich collection of $J'$ novel documents that exhibit broad topical and stylistic coverage pertinent to the target domain. Formally, the final synthetic corpus is the union of these anchor-conditioned subsets:
\[
\Dsynth = \bigcup_{i=1}^{A} \{d'_{i,1}, \ldots, d'_{i,J'_i}\}, \quad \text{where } \sum_{i=1}^{A} J'_i = J'.
\]
This hierarchical approach ensures explicit semantic coherence and comprehensive topical coverage in the resulting synthetic corpus, potentially improving the effectiveness of downstream contextual embedding adaptation. Specific implementation details on this generation process and prompting strategies are given in \cref{sec:experiments} and \cref{app:prompts}, respectively. Because $\Dsynth$ is reused verbatim during deployment, this synthesis step must only be executed once per domain, making it computationally efficient for practical application.

\subsection{Online: Inference with Synthetic Context}
During the online inference phase, we utilize the pretrained context-aware model components ($M_1$ and $M_2$ with frozen weights) and the generated synthetic context $\Dsynth$.

\paragraph{Context Embedding Precomputation.} 
At inference we feed the cached synthetic context into $M_{2}$, incurring no extra per-query overhead beyond a standard forward pass; the costly synthesis step occurs only \emph{once}. Since $\Dsynth$ is fixed, we precompute the first-stage vectors, denoting \[ \mC_{j}=M_{1}(d'_{j}), \quad d'_{j} \in \Dsynth, \] and store the set $\{\mC_{1},\dots,\mC_{J'}\}$ for reuse.

\paragraph{Final Embeddings.} 
When a new document $d$ or query $q$ arrives, we first obtain its token
embeddings,
$E(d)=\{E(w_{1}),\dots,E(w_{T})\}$ and
$E(q)=\{E(q_{1}),\dots,E(q_{T'})\}$.
We then use the second-stage encoder $M_{2}$ alongside the cached synthetic context to produce the final zero-shot contextualized embedding:
\begin{align}
\phi(d;\Dsynth) &=
   M_{2} \bigl(\mC_{1},\dots,\mC_{J'},E(d)\bigr),
   \label{eq:zs_cde_phi_d}\\
\psi(q;\Dsynth) &=
   M_{2} \bigl(\mC_{1},\dots,\mC_{J'},E(q)\bigr).
   \label{eq:zs_cde_psi_q}
\end{align}
These final embeddings incorporate domain signals derived from the synthetic context and are subsequently used for downstream retrieval tasks. Notably, the dominant
runtime cost remains the single forward pass through $M_{2}$.

%%%%%%%%%%%%%%%%%%%%%%%%%%%%%%%%%%%%%%%%%%%%%%%%%%%%%%%%%%%%%%%%%

\section{Experimental Setup}
\label{sec:experiments}

This section details the experimental protocol designed to evaluate the effectiveness of \ModelName{} in realistic zero-shot domain adaptation scenarios using established retrieval benchmarks.

\paragraph{Datasets and Metrics.} We evaluate our approach on the widely used MTEB~\citep{muennighoff2022mteb} benchmark, which covers a diverse range of embedding tasks. 
We sample $\Dex$ from the BEIR~\citep{thakur2021beir} benchmark; specifically, from those tasks that are the closest match to the domain of the target corpus (see \cref{app:exemplar_sampling} for details).
Should the same task be present across both datasets, then we choose the next most relevant one instead.
For example, for ArguAna~\citep{boteva2016}, which is present in both benchmarks, we choose the most similar task from BEIR instead to sample $\Dex$ from.
We randomly sample documents that have \(\geq 100\) tokens to provide sufficient content for the LLM to capture domain characteristics. 
This simulates a realistic scenario where a user provides a few characteristic examples for domain adaptation. 
We ensure no leakage by replacing any document that has a 20-token span overlap between $\Dex$ and the corresponding MTEB evaluation datasets.
Retrieval quality is evaluated using the standard NDCG@10 metric.

\paragraph{Baselines for Comparison.}
We compare \ModelName{} against key baselines of similar size to contextualize its performance. We establish context-agnostic performance using strong, standard biencoder models: \texttt{gte-base-en-v1.5} (GTE v1.5)~\citep{li2023gte} and \texttt{bge-base-en-v1.5} (BGE v1.5)~\citep{xiao2024bge}.
For experiments involving context-aware embeddings, we utilize the publicly available \texttt{cde-small-v1} model~\citep{morris2024contextual}. This model comprises 137M parameters and was pretrained on a large, diverse corpus. We use its original frozen weights throughout all experiments, ensuring fair comparison and isolating the effect of the context source. For this baseline, context embeddings are computed from $J=512$ real documents randomly sampled from the target corpus partition, serving as a practical upper bound using real context with a comparable context size.
Additionally, we include the Generic Synthetic Context (GSC) baseline, which generates synthetic documents using a generic prompt applied to the same LLM as \ModelName{}, but without its hierarchical approach. This baseline isolates the impact of \ModelName{}'s use of domain anchors, providing a direct comparison to a simpler synthetic context generation method.
Finally, to test the sensitivity of our approach to exemplar document selection, we include a random baseline. Here, we randomly sample 512  documents from the first 10k entries of the Colossal Clean Crawled Corpus (C4) dataset as contextual documents.
Our baseline selection directly tests \ModelName{}'s core hypothesis: using synthetic context as a drop-in replacement for real context with a frozen architecture. Consequently, methods requiring training-time adaptation (e.g., GPL~\citep{wang2022gpl}) or online corpus access (e.g., pseudo-relevance feedback~\citep{rocchio1971prf, li2018nprf}) are considered orthogonal to this specific evaluation.

\paragraph{Synthetic Context Generation.}
We generate the synthetic context $\Dsynth$ for \ModelName{} using GPT-4o (\texttt{gpt-4o-2024-11-20}) via its API, chosen for its strong instruction-following, ability to capture nuanced stylistic and topical patterns from limited examples across diverse domains, and cost-effectiveness. A carefully constructed prompt (see Appendix \ref{app:prompt_expansion}) first provides the $k=5$ curated domain examples and instructs the LLM to generate synthetic documents hierarchically per \cref{sec:gen}. This prompt is designed for outputs that are stylistically and topically aligned with these examples -- building on domain anchors -- and sufficiently diverse to form a rich context. For each of $A=20$ anchor documents $a_i \in \mathcal{D}_{\text{anchor}}$, the LLM generates an equal fraction ($J'/A$) of the $J'=512$ total synthetic documents forming $\Dsynth$. This per-anchor generation, designed to elaborate on and diversify $a_i$'s theme, proceeds in \textit{parallel}. We use default API sampling parameters for reproducibility.

\paragraph{Additional Implementation Details.} 
For \ModelName{}, the synthetic context embeddings ($\mC_j$) were pre-computed from $\Dsynth$ using a batch size of 16. 
During MTEB evaluation runs the models processed task queries and documents with a batch size of 512. 
Following the methodology of CDE, task-specific prefixes (see \cref{app:prefixes}) were applied to inputs before being processed. This ensures consistency in how the model receives data for both real-context and synthetic-context scenarios.
Experiments were conducted using NVIDIA A100 GPUs.

%%%%%%%%%%%%%%%%%%%%%%%%%%%%%%%%%%%%%%%%%%%%%%%%%%

\begin{table*}[t]
\small
\centering
\begin{tabular}{l@{\hspace{3em}}c@{\hspace{2.5em}}c@{\hspace{3em}}c@{\hspace{4em}}c@{\hspace{3em}}c@{\hspace{2em}}c}
\toprule
Task Category & \textbf{GTE v1.5} & \textbf{BGE v1.5} & \textbf{CDE} & \textbf{GSC} & \textbf{Random} & \textbf{ZEST (ours)} \\
\midrule
Classification       & 77.2 & 74.7 & 82.5 & 81.8 & 80.4 & 82.2 \\
Clustering      & 46.8 & 45.3 & 49.3 & 48.7 & 46.9 & 49.1 \\
Pair Classification          & 85.2 & 85.7 & 87.5 & 87.0 & 85.6 & 87.2 \\
Reranking       & 57.7 & 58.3 & 60.0 & 59.4 & 57.3 & 59.7 \\
Retrieval   & 54.1 & 52.8 & 55.2 & 54.6 & 52.8 & 55.0 \\
STS   & 82.0 & 81.6 & 83.3 & 82.7 & 81.4 & 83.0 \\
Summarization   & 31.2 & 30.8 & 32.7 & 32.1 & 30.4 & 32.3 \\
\midrule
Average       & 62.03 & 61.31 & \textbf{64.36} & 63.76 & 62.11 & \underline{64.07} \\
\bottomrule
\end{tabular}
\caption{Retrieval performance on the MTEB benchmark, shown across its task categories. Baselines include context-agnostic models (GTE v1.5, BGE v1.5), CDE with real context ($J=512$), ZEST with random documents, and our synthetic GSC baseline. ZEST uses $k=5$ examples and $J'=512$ synthetic documents, without accessing the target corpus. Best overall result in \textbf{bold}, best zero-shot (corpus-inaccessible) result \underline{underlined}.}
\label{tab:main_results}
\end{table*}

\section{Results and Discussion}
\label{sec:results}

This section presents the empirical evaluation of \ModelName{}, demonstrating its ability to achieve effective contextual adaptation in zero-shot scenarios. We analyze its performance against established baselines, investigate the impact of key hyperparameters through ablation studies, and discuss the implications of our findings.

\subsection{Main Results: Zero-Shot Contextual Adaptation}

The results presented in \Cref{tab:main_results} compellingly demonstrate the efficacy of \ModelName{}. Across the MTEB benchmark, \ModelName{} using its exemplar-guided synthetic context achieves performance strikingly close to the CDE model that leverages full target corpus access, with an average difference of merely 0.29 NDCG@10 points. This indicates that \ModelName{} comes within 0.45\% of the performance attainable with unrestricted access to the real corpus -- a significant finding given its zero-shot nature. While the real-context baseline naturally sets a practical upper bound, \ModelName{} closes a large portion of the gap between the no-context baseline and this upper bound. Additionally, the random baseline sets the lower bound, with a significant decrease in performance. This indicates that while our method is not particularly sensitive when choosing documents within the general domain of the target corpus, choosing exemplar documents from an unrelated domain hinders the contextual embeddings, as expected.

Notably, \ModelName{} also outperforms the GSC synthetic baseline by 0.31 NDCG@10 points on average, highlighting the benefit of our domain-anchor-based synthesis approach over simpler synthetic generation from the $k$ exemplars. Furthermore, \ModelName{} establishes substantial gains over strong context-agnostic baselines. It surpasses GTE v1.5 by an average of 2.04 and BGE v1.5 by 2.76 as measured by NDCG@10. These improvements underscore the value of carefully generated synthetic context. Indeed, by recovering 87.6\% of the performance gap between GTE v1.5 and the full-access CDE model, \ModelName{} effectively emulates the benefits of real corpus statistics without requiring direct access.

\subsection{Ablation Studies}
\label{sec:ablation}

\begin{figure}[t]
    \centering
    \includegraphics[width=0.95\linewidth]{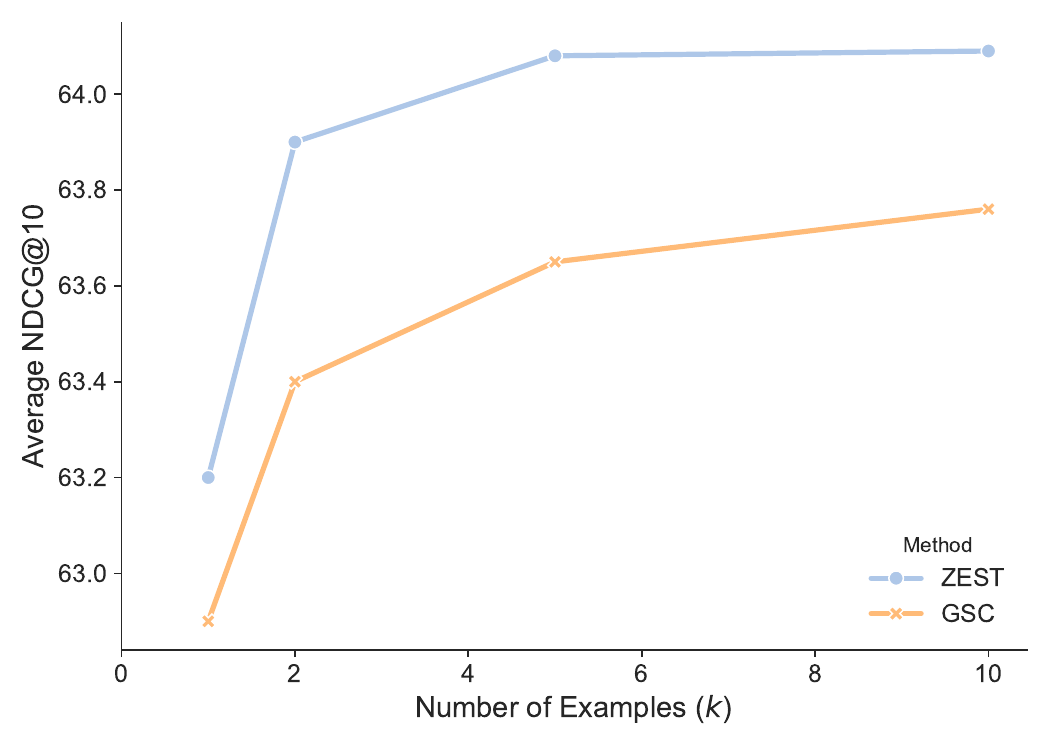}
    \caption{Performance across benchmark datasets when altering the number of few-shot examples ($k$), with constant $J'=512$ synthetic documents. We compare it to the performance of GSC, which does not use }
    \label{fig:ablation_k}
\end{figure}

\paragraph{Effect of Number of Examples and Anchors.} 
To better understand the behavior and robustness of \ModelName{}, we perform ablation studies on the number of guiding examples $k \in \{1, 2, 5, 10\}$, holding the synthetic context size fixed at $J'=512$. Results are shown in \Cref{fig:ablation_k}. Performance increases when moving from $k=1$ to $k=5$, indicating that providing the LLM with a few diverse examples significantly helps it capture the target domain's characteristics more accurately by not overfitting to a single example. Using $k=10$ provides only marginal, if any, additional benefit over $k=5$ in our experiments, suggesting that our hierarchical generation using $A=20$ domain anchors (as described in \cref{sec:gen}) provides sufficient diversity with only $k=5$ examples. Furthermore, we also investigate the number of $A$ and find that performance is not sensitive to this parameter, as variations around our default yield negligible differences in overall MTEB scores.

\begin{figure}[t]
    \centering
    \includegraphics[width=0.95\linewidth]{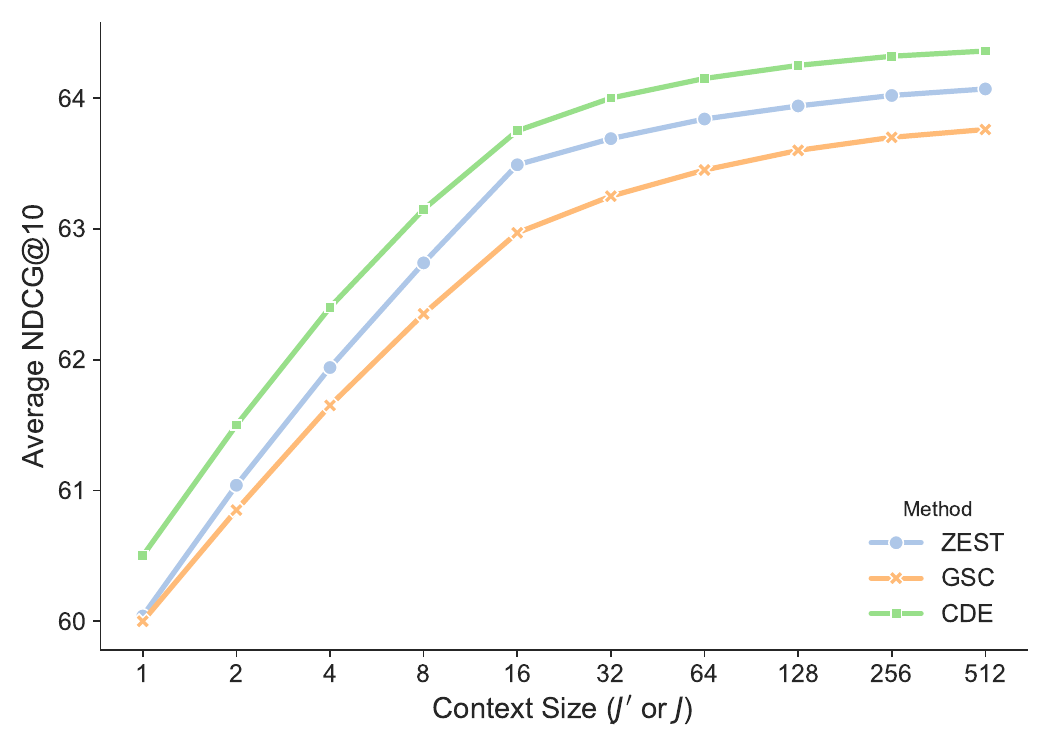}
    \caption{Performance across benchmark datasets, comparing \ModelName{} to synthetic baseline and CDE at equal context sizes, with constant $k=5$ examples.}
    \label{fig:ablation_j_prime}
\end{figure}

\paragraph{Effect of Synthetic Context Size.} We further investigate changing the context size while keeping $k$ constant. \Cref{fig:ablation_j_prime} illustrates how retrieval performance varies with $J'$. Here, performance generally improves substantially as $J'$ increases from small values (e.g., 2) towards $J'=512$. While the most significant gains often occur with initial increases in context size (improvement slows after $J'=16$), the results suggest that a larger synthetic context allows the LLM to generate a richer and more diverse set of documents, providing a more comprehensive contextual signal and leading to performance levels comparable to using real context. This justifies our choice of $J'=512$ for the main results, although smaller values of $J'$ still offer improvements.

\paragraph{Impact of LLM Choice.}
To assess the sensitivity of our synthetic context generation approach to the choice of LLM, we compared the performance when using GPT-4o versus the open-source \texttt{Llama-3.3-70B-Instruct} (Llama 70B) model. This comparison, the results of which are shown in \cref{tab:llm}, focuses on the two embedding methods directly employing LLM-generated context: our proposed \ModelName{} framework and the synthetic GSC baseline.

As shown, the advanced capabilities of GPT-4o translate to more effective synthetic context for both GSC and \ModelName{}, outperforming Llama 70B by 1.05 points for GSC and 1.11 points for \ModelName{}. However, it is noteworthy that \ModelName{} still demonstrates a clear advantage over GSC even when both utilize Llama 70B, with \ModelName{} achieving a score of 62.96, maintaining a lead of 0.25 points over GSC's 62.71. This suggests that while the quality of the generator LLM is impactful, \ModelName{}'s exemplar-guided hierarchical synthesis strategy provides inherent benefits in creating more effective domain-specific context, regardless of the specific LLM employed.

\subsection{Discussion}
\label{sec:discussion}
The strong performance of \ModelName{}, achieving comparability with CDE using real context ($J=512$), underscores the capability of modern LLMs to act as effective simulators of domain-specific corpus characteristics based on minimal examples. Effectively, we treat the LLM as a giant database that we retrieve our synthetic documents from. The synthetic context $\Dsynth$ captures not just topical relevance but also implicit statistical patterns (such as term co-occurrence and relative frequency) and stylistic elements that the pretrained CDE model leverages for adaptation. While not a perfect replacement for the real corpus, the synthetic context provides a remarkably effective proxy, enabling high-performance context-aware retrieval where it was previously infeasible.

\paragraph{Computational Considerations:} \ModelName{} introduces an offline cost for generating $\Dsynth$. Importantly, this represents a one-time, offline process per target domain. Specifically, generating $\Dsynth$ using GPT-4o via its API results in synthetic documents with an average length of 255.6 output tokens, compared to 225.1 tokens for the benchmark documents. The online inference cost for \ModelName{} involves only the standard forward pass using the precomputed synthetic context embeddings $\{\mC_j\}$, similar to that of CDE using real context. In our experiments, the time for the hierarchical generation process is negligible compared to inference. This compares favorably to the resources required for alternative adaptation strategies, which typically demand significant GPU hours for finetuning and/or assume the feasibility of corpus access for data acquisition and processing.

\paragraph{Qualitative Insights:} Examining samples from $\Dsynth$ -- shown in \cref{app:synth_samples_pipeline} -- reveals the ability of LLMs to generate relevant and stylistically consistent documents for a specific domain. However, occasional generic or less relevant documents do occur, potentially limiting performance.

\begin{table}[t]
\small
\centering
\begin{tabular}{lcc}
\toprule
LLM used for Synthesis & \textbf{GSC} & \textbf{\ModelName{}} \\
\midrule
GPT-4o & 63.76 & 64.07 \\
Llama-3.3-70B-Instruct & 62.71 & 62.96 \\
\bottomrule
\end{tabular}
\caption{Influence of the generating LLM on retrieval performance via synthetic context. Average MTEB NDCG@10 scores for GSC and \ModelName{} using $k=5$ and $J'=512$.}
\label{tab:llm}
\end{table}

%%%%%%%%%%%%%%%%%%%%%%%%%%%%%%%%%%%%%%%%%%%%%%%%%%%%%%%%%%%%%
% related_work.tex
\section{Related Work}
\label{sec:related_work}

\paragraph{Neural Dense Retrieval.}
Neural retrievers learn to map documents and queries into a shared embedding space, optimizing a contrastive loss for efficient retrieval~\citep{karpukhin2020dpr,reimers-2019-sentence-bert,ni2021gtr,izacard2022unsupervised,xiong2020ance}.  
However, their \emph{context-agnostic} design makes them vulnerable to domain shifts when the test corpus differs from pretraining data~\citep{thakur2021beir}.  
Our work builds on this foundation but injects domain signals at inference without requiring access to the full corpus.

\paragraph{Context-Aware Embeddings.}
To mitigate corpus shift, recent methods augment models with contextual information drawn from \emph{neighbor} documents.  
Contextual Document Embeddings~\citep{morris2024contextual} and late-interaction models like ColBERT~\citep{khattab2020colbert,santhanam2022colbertv2} refine each embedding by reading a subset of the target corpus at inference.  
While effective, these approaches assume unrestricted access to the entire document collection—an assumption that fails in privacy-sensitive or large-scale environments. Semi‑parametric language models such as $k$NN‑LM~\citep{khandelwal2020knnlm} or non‑parametric transformers~\citep{kossen2022selfattention} also incorporate external information, yet focus on embeddings for text generation rather than retrieval embeddings. 
In contrast, \ModelName{} replaces the real neighbor set with a compact, LLM-generated proxy, enabling analogous context-aware gains under a zero-access constraint.

\paragraph{Test-time and unsupervised retrieval adaptation.}
A rich line of work explores adapting retrievers to new domains without labeled data. Unsupervised corpus-aware pretraining methods (e.g., GPL~\citep{wang2022gpl}, LaPraDoR~\citep{xu2022laprador}, SimLM~\citep{wang2023simlm}) fine-tune on the target documents themselves, while few-shot or parameter-efficient schemes (e.g., TSDAE~\citep{wang2021tsdae}, adapters~\citep{houlsby2019adaptertuning}, prompt tuning~\citep{dai2022promptagator}) require some labeled examples.  
Test-time techniques such as pseudo-relevance feedback~\citep{rocchio1971prf,wang2021colbertprf} or Boot\&Switch~\citep{jiang2023bootswitch} adjust queries or model parameters on the fly but likewise need access or online optimization.  
Unlike these methods, \ModelName{} performs \emph{training-free} adaptation: it freezes all model weights and synthesizes a proxy context once offline from only a handful of example documents.

\paragraph{Synthetic data generation for retrieval.}  
LLMs are now routinely used in retrieval-related tasks to fabricate documents via few-shot prompting~\citep{10.1145/3477495.3531863}, for supervision -- such as with Promptagator~\citep{dai2022promptagator} and  CRAFT~\citep{ziegler2024craft} -- or to curate hard negatives~\citep{solatorio2024gistembedguidedinsampleselection}.  Prior work typically deploys the synthetic text for \emph{training}~\citep{shao2025reasonirtrainingretrieversreasoning}.  In contrast, \ModelName{} exploits LLMs at \emph{inference time}: the generated mini‑corpus acts as a stand‑in reference that unlocks corpus‑aware embeddings without accessing the target corpus.

%%%%%%%%%%%%%%%%%%%%%%%%%%%%%%%%%%%%%%%%%%%%%%%%%%%%

\section{Conclusion}
We introduced \ModelName{}, a novel method that enables context-aware document retrieval adaptation without requiring access to the target corpus. By employing LLMs guided by few-shot examples to synthesize a representative context corpus offline, \ModelName{} allows a pretrained contextual embedding model to adapt effectively during online inference. Our findings show \ModelName{} substantially improves zero-shot retrieval over context-agnostic methods, nearing the performance of models with full corpus access. \ModelName{} addresses a critical practical limitation of context-aware models, offering a viable path toward more adaptable and effective document embeddings in scenarios constrained by corpus access, privacy, or scale. This data-centric adaptation strategy opens new possibilities for deploying sophisticated retrieval models in challenging real-world environments.

\section*{Limitations}
Despite its effectiveness, \ModelName{} has limitations. Its performance is inherently linked to the quality of the LLM-generated synthetic context, as it remains a proxy for the true corpus statistics. The reliance on LLMs introduces dependencies on external APIs and associated costs for the offline generation step. Furthermore, potential biases present in the LLM or $\Dex$ could be amplified in the synthetic context, requiring careful consideration in sensitive applications. Finally, the selection of $k$ examples introduces variability; automating or guiding this selection could improve robustness. We leave this to future work.

Future work could further explore: (1) finetuning open-source LLMs specifically for high-fidelity context generation to reduce dependencies and (2) developing techniques for automated quality assessment of the generated $\Dsynth$ and potentially filtering or refining it.
%%%%%%%%%%%%%%%%%%%%%%%%%%%%%%%%%%%%%%%%%%%%%%%%%%%%%%%%

\bibliography{acl_latex}

\appendix

\section{Prompting Strategies for Synthetic Corpus Generation}
\label{app:prompts}
This section details the prompting strategies we employ to generate the synthetic context corpus $\Dsynth$. The process, as described in \Cref{sec:gen}, is hierarchical, involving two main steps: (1) Domain Anchor Generation and (2) Synthetic Corpus Expansion. We used the default API sampling parameters (e.g., temperature) for all generations to ensure reproducibility. The $k=5$ exemplar documents $\Dex$ are assumed to be provided as part of the input to the LLM for the first step.

\paragraph{Step 1: Domain Anchor Generation}
\label{app:prompt_anchors}

The objective of this step is to generate $A$ diverse domain anchor documents from the exemplar set $\Dex$ to create $\mathcal{D}_{\text{anchors}}$. These anchors serve as thematic seeds. They are generated sequentially to encourage diversity and avoid thematic repetition. For each anchor $a_i$, the LLM is instructed to produce a concise document that captures a distinct topical or stylistic facet present in the provided exemplars, while also being mindful of previously generated anchors in the sequence (if applicable).
The generalized prompt structure for generating a single domain anchor $a_i$ is shown in Figure~\ref{fig:domain_anchor_prompt}.

\paragraph{Step 2: Synthetic Corpus Expansion}
\label{app:prompt_expansion}

Once the domain anchors are generated, the synthetic corpus $\Dsynth$ (of size $J'$) is created by expanding upon these anchors. For each anchor document $a_i \in \mathcal{D}_{\text{anchors}}$, the LLM is prompted to generate ($J'/A$) of synthetic documents (we assign one extra document per anchor until the remainder is exhausted). This step is performed in parallel for each anchor. The goal is for the LLM to elaborate on and diversify the theme encapsulated by the specific anchor $a_i$, producing a set of full-length, representative documents.
The generalized prompt structure for expanding a single domain anchor is shown in Figure~\ref{fig:synthetic_corpus_prompt}.

\section{Details on Exemplar Set Sampling}
\label{app:exemplar_sampling}

To enable zero-shot contextual adaptation in \ModelName{}, we rely on a small exemplar set \(\Dex = \{d_{\text{ex}}^{1}, \ldots, d_{\text{ex}}^{k}\}\) to guide the generation of the synthetic context corpus \(\Dsynth\). As described in \Cref{sec:experiments}, these exemplars are sourced from the BEIR benchmark. This section details the process of selecting \(\Dex\), the mapping of BEIR tasks to unique domain keywords, and the measures taken to ensure no information leakage between \(\Dex\) and the MTEB evaluation datasets.

\paragraph{Mapping BEIR Tasks to Domain Keywords.}
To systematically select exemplars that typify the domain of an MTEB target task, we first assign each of the 18 BEIR tasks a unique keyword that encapsulates its primary domain or task characteristic. These keywords serve as an intermediary representation, allowing us to later align MTEB tasks with the most relevant BEIR-derived exemplars based on domain similarity. \Cref{tab:beir_keyword_mapping} presents this mapping, with each keyword chosen to be distinct and representative of the task’s content.

The keyword assignment prioritizes the dominant domain or retrieval objective of each BEIR task. For instance, biomedical tasks like BioASQ and TREC-COVID are assigned keywords like ``BiomedQA'' and ``COVIDResearch,'' respectively, to distinguish their focus within the broader biomedical domain. Similarly, tasks like Quora and CQADupStack, both involving question answering, are differentiated by keywords ``DuplicateQA'' and ``ForumQA,'' reflecting their specific contexts (duplicate question detection versus forum-based Q\&A). This approach ensures that the keywords are sufficiently granular to avoid overlap while remaining general enough to facilitate alignment with MTEB tasks. For MTEB tasks, we select the BEIR task whose keyword best matches the MTEB task’s domain or retrieval goal, determined by manual inspection of task descriptions and data characteristics. If an MTEB task corresponds to a BEIR task (e.g., ArguAna), we select the next closest task to avoid direct overlap, as noted in \Cref{sec:experiments}.

\paragraph{Practical Considerations.}
The use of BEIR tasks as a source for \(\Dex\) leverages their diversity and public availability, making the approach reproducible and scalable. The keyword-based mapping simplifies the alignment of MTEB tasks to appropriate exemplars while avoiding direct task-to-task dependencies, which could risk evaluation bias. By sampling only a small number of documents, we simulate a realistic scenario where practitioners provide minimal domain examples, aligning with \ModelName{}’s goal of minimal input requirements. The ablation studies in \Cref{sec:ablation} confirm that our approach provides sufficient diversity for effective synthetic context generation, supporting the robustness of this sampling strategy.

% Table mapping BEIR tasks to keywords
\begin{table}[t]
\small
\centering
\begin{tabular}{l@{\hspace{4em}}l}
\toprule
\textbf{BEIR Task} & \textbf{Unique Keyword} \\
\midrule
MS MARCO & WebSearch \\
TREC-COVID & COVIDResearch \\
NFCorpus & Nutrition \\
BioASQ & BiomedQA \\
HotpotQA & MultiHopQA \\
FiQA-2018 & FinanceQA \\
Signal-1M (RT) & SocialMedia \\
TREC-NEWS & NewsSearch \\
Robust04 & NewsArchive \\
ArguAna & Argumentation \\
Touché-2020 & Debate \\
CQADupStack & ForumQA \\
Quora & DuplicateQA \\
DBPedia-Entity & EntityRetrieval \\
SCIDOCS & Citation \\
SciFact & SciFactCheck \\
Climate-FEVER & ClimateClaims \\
FEVER & FactCheck \\
\bottomrule
\end{tabular}
\caption{Mapping of BEIR tasks to unique domain keywords. Each keyword encapsulates the primary domain or task characteristic, enabling alignment with MTEB tasks based on domain similarity.}
\label{tab:beir_keyword_mapping}
\end{table}

\section{Task-Specific Prefixes}
\label{app:prefixes}
We use standard prefixes, hand-written for each MTEB evaluation dataset, across all our evaluations. The prefix selection procedure follows the methodology outlined in~\cite{nussbaum2025nomicembedtrainingreproducible}. The specific prefix categories are:
\begin{itemize}
    \item \texttt{Search query}
    \item \texttt{Search document}
    \item \texttt{Classification}
    \item \texttt{Clustering}
\end{itemize}

Using these prefixes helps the model identify the task at hand and ensures consistency in how the model receives data for both real-context and synthetic-context scenarios.

\section{Examples of Generated Synthetic Documents: Full Pipeline}
\label{app:synth_samples_pipeline} % Label for this new, detailed section

This section provides examples illustrating the full pipeline used by \ModelName{} to generate synthetic documents for the $\Dsynth$ corpus. These examples demonstrate how an initial exemplar document ($\Dex$) from a specific domain guides the generation of a domain anchor, which in turn seeds the creation of a final synthetic document. This hierarchical process, as described in \Cref{sec:gen}, is based on $k$ exemplar documents (for clarity, we show the pipeline for $k=1$ exemplar in each domain example below, referenced as examples in \Cref{artifact:biomed_dex} through \Cref{artifact:financial_synth}).

\paragraph{Example 1: Biomedical Domain}
An exemplar document focusing on genetic recoding in Archaea (see \Cref{artifact:biomed_dex}) was provided to ground the generation process in the biomedical domain. This initial document serves as the primary input for the LLM to understand the target domain's characteristics.

Based on the provided biomedical exemplar, the LLM generated a domain anchor (see \Cref{artifact:biomed_anchor}). This anchor encapsulates a core theme derived from the exemplar -- in this case, programmed ribosomal frameshifting in Archaea -- and serves as a more focused seed for subsequent document generation.

Expanding upon the biomedical domain anchor, the LLM then produced a full synthetic document (see \Cref{artifact:biomed_synth}). This final document elaborates on the mechanisms and implications of frameshifting, demonstrating how the anchor guides the creation of a more detailed and contextually relevant piece of text for the synthetic corpus.

\paragraph{Example 2: Financial Domain}

For the financial domain, an exemplar document discussing interest rates and loan types (see \Cref{artifact:financial_dex}) was used, which leads to a corresponding domain anchor (see \Cref{artifact:financial_anchor}), and, finally, a synthetic document (see \Cref{artifact:financial_synth}). This document also explores lender strategies and borrower behavior in response to varying interest rate environments.

%%%%%%%%%%%%%%%%%%% APPENDIX ARTIFACTS %%%%%%%%%%%%%%%%%%%
\onecolumn

\begin{appendixartifact}{Prompt for Domain Anchor Generation}
\begin{lstlisting}[style=mypromptstyle]
Systematically examine the {k} exemplar documents provided below to extract and synthesize 
their core themes, stylistic patterns, and domain-specific terminology. Leverage this 
analysis to craft a new domain anchor document that encapsulates these elements.

Here are the exemplar documents:

Exemplar 1:
"""
{exemplar_document_1_text}
"""

Exemplar 2:
"""
{exemplar_document_2_text}
"""
...

Exemplar k:
"""
{exemplar_document_k_text}
"""

Previously generated anchor documents (if any):
- {anchor_1}
- {anchor_2}
- {anchor_i-1}


Your task is to generate a new, concise domain anchor document. This document should:
1. Be approximately as long as the exemplar documents.
2. Capture a distinct and specific topical theme, concept, or stylistic characteristic
   evident in the exemplar documents.
3. Cover key terminology, entities, and typical writing style of the domain as
   represented by the exemplars.
4. If previous anchors were mentioned, ensure this new anchor explores a
   DIFFERENT facet or theme than those already covered to maximize diversity.
5. The anchor should be a coherent piece of text, similar to the exemplar documents, 
   not just a list of keywords.

Generate only the domain anchor document itself.
\end{lstlisting}
\end{appendixartifact}
\begin{center}
    \captionof{figure}{Prompt for Domain Anchor Generation}
    \label{fig:domain_anchor_prompt}
\end{center}

\vspace{1em} % Add some vertical space between artifacts

\begin{appendixartifact}{Prompt for Synthetic Corpus Expansion}
\begin{lstlisting}[style=mypromptstyle]
You are tasked with generating a document that is representative of a specific 
domain and theme.

You are given the following domain anchor document to build on, which encapsulates 
a key theme or stylistic element of the target domain:

Domain Anchor:
"""
{domain_anchor_document_text}
"""

Your task is to generate another full synthetic document that elaborates on,
exemplifies, and diversifies the core theme and style presented in the domain anchor.
This new document should:
1. Be topically coherent with the provided domain anchor.
2. Be a complete, well-structured document (e.g., an article, a report excerpt,
   a descriptive passage) of similar length.
3. Should explore various sub-topics, perspectives, or aspects related to the main 
   theme of the anchor, ensuring diversity among them.
4. Maintain a style (e.g., tone, vocabulary, sentence structure) consistent with the
   domain anchor and typical of the implied domain.
5. Be factually plausible and internally consistent, even if entirely synthetic.

Respond only with your generated document. Ensure the document is clearly separated
by placing "---DOCUMENT END---" at the end of the document you generate.
\end{lstlisting}
\end{appendixartifact}
\begin{center}
    \captionof{figure}{Prompt for Synthetic Corpus Expansion}
    \label{fig:synthetic_corpus_prompt}
\end{center}

\vspace{1em} % Add some vertical space between artifacts

\begin{appendixartifact}{Biomedical Exemplar Document}
\begin{lstlisting}[style=mycodestyle]
The standard rules of genetic translational decoding are altered in specific genes 
by different events that are globally termed recoding. In Archaea recoding has been 
unequivocally determined so far only for termination codon readthrough events. We 
study here the mechanism of expression of a gene encoding for a $\alpha$-l-fucosidase 
from the archaeon Sulfolobus solfataricus (fucA1), which is split in two open reading
frames separated by a -1 frameshifting. The expression in Escherichia coli of the 
wild-type split gene led to the production by frameshifting of full-length polypeptides 
with an efficiency of 5%. Mutations in the regulatory site where the shift takes place
demonstrate that the expression in vivo occurs in a programmed way. Further, we 
identify a full-length product of fucA1 in S.solfataricus extracts, which translate
this gene in vitro by following programmed -1 frameshifting. This is the first
experimental demonstration that this kind of recoding is present in Archaea.
\end{lstlisting}
\end{appendixartifact}
\begin{center}
    \captionof{figure}{Example of an input exemplar document ($\Dex$) for the biomedical domain, detailing genetic recoding in Archaea. This document serves as the initial seed for the LLM.}
    \label{artifact:biomed_dex}
\end{center}

\vspace{1em}

\begin{appendixartifact}{Biomedical Domain Anchor}
\begin{lstlisting}[style=mycodestyle]
Programmed ribosomal frameshifting represents a crucial gene expression regulation
mechanism in Archaea, enabling the synthesis of alternative protein products from a
single mRNA. This recoding event, often involving a -1 shift, allows organisms like
*Sulfolobus solfataricus* to expand their proteomic diversity and adapt to specific
environmental conditions. Understanding the cis-acting RNA elements and trans-acting
factors that govern frameshift efficiency in these extremophiles is key to deciphering
their unique biology and potential biotechnological applications in areas such as
novel enzyme discovery or controlled protein synthesis. The study of such mechanisms
also sheds light on the evolutionary plasticity of the genetic code.
\end{lstlisting}
\end{appendixartifact}
\begin{center}
    \captionof{figure}{Domain anchor generated from the biomedical exemplar (see Figure~\ref{artifact:biomed_dex}). It focuses on the theme of programmed ribosomal frameshifting in Archaea.}
    \label{artifact:biomed_anchor}
\end{center}

\vspace{1em}

\begin{appendixartifact}{Biomedical Synthetic Document}
\label{artifact:biomed_synth}
\begin{lstlisting}[style=mycodestyle]
The efficiency of programmed -1 ribosomal frameshifting in archaeal organisms, such as the
crenarchaeon *Sulfolobus solfataricus*, is intricately regulated by specific mRNA sequences
and structural motifs located proximal to the frameshift site. These regulatory elements
typically include a 'slippery sequence' (often a heptanucleotide like A AAA AAC) where the
ribosome can lose its grip on the mRNA, and a downstream RNA secondary structure, commonly
a pseudoknot or a stable hairpin. This structure is thought to induce ribosomal pausing,
providing a kinetic window for the realignment of the tRNA in the P-site to the -1 frame.
The interplay between the ribosome's own components, the mRNA transcript, various
translation factors, and the nascent peptide chain within the exit tunnel collectively
influences the probability and fidelity of a successful frameshift event. Current research
endeavors focus on genome-wide mapping of these frameshift-prone sites in diverse Archaea,
aiming to uncover novel regulatory networks and identify potential targets for antimicrobial
agents or innovative tools for synthetic biology, such as precisely controlling the
stoichiometric production of multiple proteins from a single engineered transcript.
\end{lstlisting}
\end{appendixartifact}
\begin{center}
    \captionof{figure}{Final synthetic document ($\Dsynth$) for the biomedical domain, expanded from the domain anchor (see Figure~\ref{artifact:biomed_anchor}), discussing the regulation and implications of ribosomal frameshifting.}
    \label{artifact:biomed_synth}
\end{center}

\vspace{1em}

\begin{appendixartifact}{Financial Exemplar Document}
\begin{lstlisting}[style=mycodestyle]
A lot of loans are taken out on a fixed rate basis, so the rate is part of the contract
and is therefore covered by contract law. If the loan is taken out on a variable basis then
in principle the rate can rise within the terms of the contract. If a particular lender
tries to raise its rates out of line with the market then its customers will seek
alternative, cheaper, loans and pay off their expensive loan if they can. If rates rise
sharply in general due to unusual politico-economic circumstances then those with variable
rate loans can find themselves in severe trouble. For example the base rate in the UK
(and therefore variable mortgage rates closely tied to it) spiked sharply in the late
80s which caused severe stress to a lot of borrowers and undoubtedly pushed some into
financial difficulties.
\end{lstlisting}
\end{appendixartifact}
\begin{center}
    \captionof{figure}{Example of an input exemplar document ($\Dex$) for the financial domain, discussing factors influencing interest rates and loan types.}
    \label{artifact:financial_dex}
\end{center}

\vspace{1em}

\begin{appendixartifact}{Financial Domain Anchor}
\begin{lstlisting}[style=mycodestyle]
The interplay between fixed and variable interest rates significantly impacts borrower
stability and broader market dynamics. Fixed-rate loans offer predictability and
insulate borrowers from sudden, adverse rate hikes, thereby fostering financial planning
certainty. Conversely, variable-rate loans, while potentially offering lower initial
rates, expose borrowers to market fluctuations, which can lead to payment shocks and
financial distress during periods of rising interest rates. Central bank monetary
policies, inflation expectations, and overall macroeconomic health are primary drivers
influencing these rates, creating a complex and evolving landscape for both lenders
managing risk and borrowers making critical financial decisions.
\end{lstlisting}
\end{appendixartifact}
\begin{center}
    \captionof{figure}{Domain anchor generated from the financial exemplar (see Figure~\ref{artifact:financial_dex}). It centers on the impact of fixed versus variable interest rates and market conditions.}
    \label{artifact:financial_anchor}
\end{center}

\vspace{1em}

\begin{appendixartifact}{Financial Synthetic Document}
\begin{lstlisting}[style=mycodestyle]
Financial institutions meticulously navigate the complexities of fluctuating interest rate 
environments by strategically managing their portfolios of fixed and variable rate loan 
products. This often involves sophisticated hedging strategies using interest rate swaps or 
other derivatives to mitigate the risk associated with rate volatility. For consumers and 
businesses, a thorough understanding of their loan agreements' terms -- particularly clauses 
related to rate adjustments -- is paramount. During periods when monetary policy tightens 
and rate increases are anticipated, there's typically a discernible surge in demand for 
refinancing variable-rate debt (like adjustable-rate mortgages or ARMs) into fixed-rate
alternatives, although eligibility for such refinancing heavily depends on the borrower's
creditworthiness and prevailing market conditions. Conversely, when central banks signal an
easing cycle and rates are expected to decline, variable-rate loans might appear more
attractive due to potentially lower initial payments. However, these carry the inherent risk
of future increases should economic conditions shift unexpectedly. Lenders also adjust their
credit scoring models and underwriting standards in response to the perceived risk in the
interest rate cycle.
\end{lstlisting}
\end{appendixartifact}
\begin{center}
    \captionof{figure}{Final synthetic document ($\Dsynth$) for the financial domain, expanded from the domain anchor (see Figure~\ref{artifact:financial_anchor}), detailing lender strategies and borrower behavior in different interest rate environments.}
    \label{artifact:financial_synth}
\end{center}

\twocolumn

\end{document}